\documentclass[1p]{elsarticle}
\usepackage{lineno,hyperref} %linenoは行番号, hyperrefは分からない
\modulolinenumbers[5] %何行ごとに行番号をふるか

\journal{arxiv}%Neural Networksに出すときはNeural Networksに変更しようかね

\usepackage{bm}
\usepackage{amsmath}
\DeclareMathOperator*{\argmax}{arg\,max}
\usepackage{amsfonts}

\usepackage[subrefformat=parens]{subcaption}

\usepackage{multirow}
\begin{document}

\begin{frontmatter}

\title{Deep Q-network using reservoir computing \\ with multi-layered readout}
% \tnotetext[mytitlenote]{}

\author{Toshitaka Matsuki}
\address{Oita University, 700 Dannoharu, Oita, JAPAN 870–1192}
\ead{matsuki@oita-u.ac.jp}

% \author[mymainaddress,mysecondaryaddress]{Elsevier Inc}

% \author[mysecondaryaddress]{Global Customer Service\corref{mycorrespondingauthor}}
% \cortext[mycorrespondingauthor]{Corresponding author}

% \address[mymainaddress]{1600 John F Kennedy Boulevard, Philadelphia}
% \address[mysecondaryaddress]{360 Park Avenue South, New York}

\begin{abstract}
Recurrent neural network (RNN) based reinforcement learning (RL) is used for learning context-dependent tasks and has also attracted attention as a method with remarkable learning performance in recent research.
 However, RNN-based RL has some issues that the learning procedures tend to be more computationally expensive, and training with backpropagation through time (BPTT) is unstable because of vanishing/exploding gradients problem. 
An approach with replay memory introducing reservoir computing has been proposed, which trains an agent without BPTT and avoids these issues.
 The basic idea of this approach is that observations from the environment are input to the reservoir network, and both the observation and the reservoir output are stored in the memory. 
This paper shows that the performance of this method improves by using a multi-layered neural network for the readout layer, which regularly consists of a single linear layer. The experimental results show that using multi-layered readout improves the learning performance of four classical control tasks that require time-series processing. 
\end{abstract}

\begin{keyword}
% \texttt{elsarticle.cls}\sep \LaTeX\sep Elsevier \sep template
% \MSC[2010] 00-01\sep  99-00
deep reinforcement learning, replay memory, echo state network, reservoir computing
\end{keyword}

\end{frontmatter}

% \linenumbers  %これがあると行番号が表示される．

\section{Introduction}
Various methods have been proposed for successfully training deeply layered neural network (NN) for many years. Recently, these techniques have been called Deep Learning (DL), and DL has been shown to enable NN to outperform human-designed systems\cite{krizhevsky2012imagenet, simonyan2014very}. DL involves supervised learning, in which the NN is given a training signal as a desired output for the input, and unsupervised learning, in which the NN finds patterns in the data.

An approach to train NN with reinforcement learning (RL) has also been studied for many years \cite{anderson1987strategy, anderson1989learning, Sutton}. In RL, the agent explores in the environment and learns based on the rewards. This approach has the advantage that NN does not need explicit training signals given by humans. Tesauro et al. showed that an NN can be trained with RL to acquire ability comparable to professional players in a backgammon game \cite{Tesauro}. Shibata et al. have successfully trained an NN, which composes the entire process from image input to output of an action selection, with RL. \cite{shibata2008learning}. Recently, Mnih et al. introduced DL techniques to this method and showed that agents could complete some Atari games with performance equal to or better than human players \cite{Mnih2013}. Since then, the approach to train NN with RL, called deep reinforcement learning (DRL), has been actively studied, and its learning performance has improved dramatically. 

In RL, the experience that an agent obtains from exploring in the environment is an essential factor for learning. However, there is a problem that the correlations among the obtained experience destabilize the learning. One of the solutions to this problem is a method called “experience replay.” In this method, the agent's experience is stored in a replay memory \cite{lin1992reinforcement}. During training, the experience data is randomly sampled from the memory, and batch training is performed. This approach combines the powerful learning performance of DL with RL and realizes to stabilize the learning. In the field of DRL, various other studies have been conducted to improve learning performance.

The real world we live in has spatio-temporal dynamics, and in such an environment, context is essential. In other words, the information observed in both the present and the past is necessary to predict future and to behave appropriately in the world. The environment in which an agent can only observe uncertain state to predict future is called a partially observable Markov decision process (POMDP). In such an environment, performing time-series processing is effective to behave appropriately.

Recurrent Neural Network (RNN) is one of the NN models that have internal feedback structure and is effective for time-series processing. Backpropagation through time (BPTT) is often used to train RNN. This method trains RNN by propagating the error signal of the outputs backward through time in the direction opposite to the flow of the time-series processing in the RNN. In BPTT, the feedback structure of the RNN is expanded in time, and the error signal is propagated by considering the feedback as a multi-layered structure with shared weight values. The weight values of the same layer are multiplied many times in BPTT; therefore, the vanishing or exploding of the gradients are often caused, and learning is destabilized \cite{bengio1994learning, pascanu2013difficulty}. Various techniques or models have been proposed to solve these problems and Long Short-Term Memory (LSTM) is widely used recently \cite{hochreiter1997long, gers2000learning, cho2014learning, zaremba2014recurrent}.

Liquid State Machine (LSM) and Echo State Network (ESN) were independently proposed by Maass and Jaeger, respectively, to avoid the problem of training RNN \cite{jaeger2001echo, maass2002real}. Although both models are essentially the same, this study focuses on ESN composed of rate model neurons. ESN has a special RNN called the reservoir, whose recurrent weight matrix is fixed and only modifies the weights of the output layer called readout to generate the network outputs from the reservoir outputs for optimization. The reservoir retains the memory of the time-series input as internal dynamics and processes them non-linearly. Since the ESN modifies only the output layer, it does not need to backward process through time like BPTT, and thus it can learn time-series processing stably and quickly.

To learn tasks that depend not only on the current state but also on the past state, RNN is effective for RL \cite{lin1992reinforcement, Lin+Mitchell:1993}. However, training an RNN using RL with experience replay has a problem. It is how to train RNN that require series information using the instantaneous experience of each time step stored in the replay memory. Hausknecht et al. proposed deep recurrent Q-network (DRQN) as an approach to such a problem. They showed that LSTM can learn with DRL methods that samples experience data keeping the time step order. \cite{hausknecht2015deep}. Since then, new ideas have been proposed, and the learning performance of RL using RNN has improved dramatically \cite{lample2017playing, kapturowski2018recurrent, moreno2019performing}. However, the algorithms of these methods tend to be more complex and computationally expensive, and the vanishing/exploding gradients problem still remains.

Chang proposed deep echo state Q-network in which the system stores the output of the ESN in replay memory to avoid the issues of large computational cost and vanishing/exploding gradients problem \cite{chang2020deep}. The basic idea of this approach is to input the observation from the environment to the reservoir and store the observations and the reservoir output containing the spatio-temporally expanded memory of the observations into the replay memory. This method allows the agent to learn without ignoring the context of the environment by simple random sampling of the experiences at each time. 

The basic readout of ESN is a single layer that performs a linear transformation of the reservoir outputs. However, this study uses a multi-layered NN as a readout. Using multi-layered readout is not common because it often gives inferior results, although a deeply layered NN is more powerful and expressive to map from input to output than a linear layer \cite{lukovsevivcius2009reservoir}.
On the other hand, several studies have shown that using a multi-layered neural network as a readout improves the learning performance of agents when ESN is used for RL \cite{bush2008modeling, matsuki2017reinforcement}. This paper shows that the learning performance of deep Q-network using replay memory with ESN is improved by introducing multi-layered readout. The experimental result shows that the improvement of performance with four classical control tasks that require time-series processing.

The remainder of this paper is organized as follows: section \ref{sec:background} describes background of this study; section \ref{sec:method} introduces the experimental method; section \ref{sec:task} explains the tasks for verification; section \ref{sec:result} presents the experimental results; section \ref{sec:conclusion} concludes this study.

\section{Background}
\label{sec:background}
\subsection{Partially observable Markov decision process}
RL is a framework in which the agent interacts with the environment and improves its policy to maximize the expected value of the total discounted reward obtained from the environment.
The environment in RL is generally assumed to be Markov property and called Markov decision process (MDP). An MDP is a discrete-time evolution in which the state transitions in the environment from time $t$ to the next time $t+1$ depends on a probability distribution determined only by the current state $s_t$ and the action $a_t$.
% \subsection{Deep reinforcement learning}
% The approach of training an NN by RL has been studied for a long time. In this context, Deep Q-Network (DQN) by Mnih et al. has brought a breakthrough \cite{Mnih2013}.
% By integrating RL with DL technology, which had been attracting much attention for its amazing performance as machine learning algorithms, DQN learned the end-to-end process from raw pixel sensor information to action selection in Atari video games, and completed some games with the same or better performance than humans. With this success, such an approach has come to be called deep Reinforcement Learning (DRL), and more active research has been conducted. 

However, there are few cases that can assume the environment to be an MDP in practice. For example, in Atari's Breakout game, it is impossible to know which way the ball is going by observing only a game screen at a moment, and the agent cannot decide which way to move the racket. The process in which Markov property cannot be assumed is called a partially observable Markov decision process (POMDP). The agent does not receive the state $s$, but instead observes the observation $o$ in the POMDP environment.

\subsection{Deep recurrent Q-network}
To treat the Atari game environments as an MDP, DQN uses $4$ frames of the game screen as input for the NN. Therefore, DQN is unable to learn tasks that require more past memories than the input frames. In other words, for DQN given $n$ frames of input, a task that requires more than $n$ frames memory becomes a POMDP. To address these issues Matthew et al. proposed and verified DRQN in which the agent network is composed of LSTM, which is the most widely used RNN model \cite{hausknecht2015deep}. Chen et al. introduced attention architecture into the DRQN approach and verified that DRQN achieve higher scores in tasks that are difficult for DQN to learn and showed that their performance is hampered in certain games \cite{chen2016deep}. In addition, various studies have been conducted to improve the learning performance of DRQN.

\subsection{R2D2}
In DRQN, the technical challenge is to train the RNN from the experiences at each time stored in the replay memory. In the work of Matthew et al., the state of the hidden layer was initialized to 0 at the first step in training, and sequence of some steps or all step in an episode was replayed. However, the former method has limitations in learning to exploit long temporal correlations, and the latter has too strong correlations between the replayed sequences.
Kapturowski et al. proposed R2D2 that introduces the ``stored state" method that stores the state of the recurrent layer and the experience, and the "burn-in" method that uses the first part of the replay sequence only for unrolling the network state to generate the starting state \cite{kapturowski2018recurrent}. In addition, R2D2 combines methods with Ape-X by Dan et al., \cite{horgan2018distributed} and it showed remarkable learning performance over previous RL methods. On the other hand, the procedures for experience sampling have become very complex and computationally expensive .

\section{Method}
\label{sec:method}
\subsection{Q-learning}
\begin{figure}
\centering
\includegraphics[width=5.0cm]{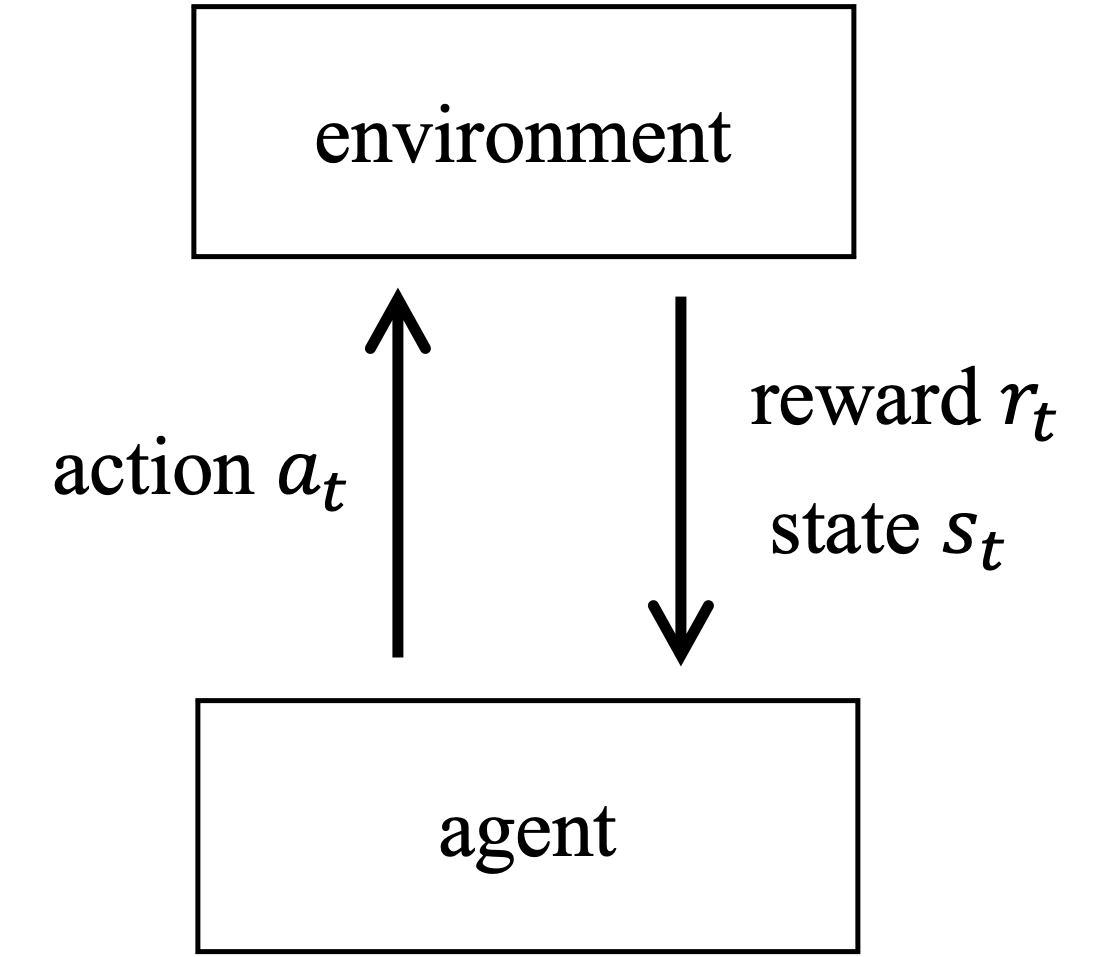}
\caption{Interaction in reinforcement learning
}
\label{fig:RL}
\end{figure}

In RL, the agent learns by interaction with their environment. Fig. \ref{fig:RL} shows the basic concept of this interaction. The agent observes the current state $s_t$ in the environment and selects an action $a_t$. The environment then transitions the current state to the next state $s'$ according to the action and informs the agent of the next state and the reward obtained there.
Q-learning is one of the most used RL methods. In Q-learning, an agent has an action-value function $Q(s,a)$, which means the value of selecting action $a \in \mathcal{A} = \{1,...,|\mathcal{A}|\}$ in state $s \in \mathcal{S}$. When an agent selects action $a_t$ in state $s_t$, transitions to state $s_{t+1}$ and obtain reward $r_{t+1}$, the appropriate value of action $a_t$ in state $s_t$ is estimated as follows
\begin{equation}
\label{eq:Q_hat}
    \hat{Q}(s_t, a_t) = r_{t+1} + \gamma \max_a Q{(s_{t+1}, a)},
\end{equation}
where $\gamma$ is a discount rate that determines the present value of discounted future rewards.
The agent learns a policy for the task by optimizing its action-value function based on these estimates while exploring in the environment. The $\epsilon$-greedy method is often used as an exploration method. In this method, the agent randomly selects an action with probability $\epsilon$ and takes the maximum value action derived by the action-value function with probability $(1-\epsilon)$.

\subsection{Methods of Deep Q-network}
In Deep Q-network, an NN that takes the state as input and outputs the estimated action-values is used as the action-value function Q. The basic deep Q-network methods, as shown below, are used in this study.

\subsubsection{Experience Replay}
Experience replay is a technique to prevent learning instabilities caused by correlations between experience data. The agent stores the experience $e_k=(s_t,a_t, r_t, s_{t+1})$ that it gained in interaction with the environment in a replay memory $\mathcal{D}=e_{k},e_{k-1},e_{k-2},\cdots$ of maximum size $N$. When the number of data in the memory exceeds $N$, the oldest data is deleted. If more than $N_D$ data are stored, the Q network is trained with randomly sampled $N_D$ data from $\mathcal{D}$. In the POMDP environment, the agent does not directly receive the state $s$, but instead observes the observation $o$, therefore the experience is stored as $e_k=(o_t,a_t, r_t, o_{t+1})$. In this study, the output $\bm{x}$ of the reservoir is additionally stored into $\mathcal{D}$.

\subsubsection{Double DQN}
Using the same values to select an action and evaluate the action value in Eq. (\ref{eq:Q_hat}) causes an overestimate of the value function. Hasselt et al. proposed Double DQN \cite{van2016deep} to prevent it. In this method, the main Q network $Q_m$, which determines the behavior in the environment, and the target Q network $Q_t$, which determines the action value during training, are implemented. The appropriate action value is estimated as follows
\begin{equation}
\label{eq:Double_DQN}
    \begin{split}
        a_m &= \argmax_a Q_m (s_{t+1}, a; \theta_i), \\
        \hat{Q}_m (s_t, a_t) &= r_{t+1} + Q_t(s_{t+1}, a_m,; \theta^-),
    \end{split}
\end{equation}
where $\theta_i$ and $\theta^-$ are the parameters of $Q_m$ and $Q_t$, respectively.
$Q_m$ is trained with Eq. \ref{eq:Double_DQN} and the data sampled from $\mathcal{D}$ at each step, and $Q_t$ is updated by copying $\theta_i$ to $\theta^-$ every two episodes.

\subsubsection{Reward clipping}
The reward for each step is limited to $-1$, $0$, or $1$. It is known that reward clipping is effective in reducing the variation of hyperparameters among tasks.

% \subsubsection{Dueling network}
% The network output is separated into $V(s;\theta, \beta)$, which depends on the state, and vector $A(s, a;\theta,\alpha)$, which depends on both the action and the state. In this experiment, the implementation of the action value was based on the following equation proposed by Wang et al. \cite{wang2016dueling} as
% \begin{equation}
% \label{eq:Dueling_Network}
%     Q(s, a;\theta, \alpha, \beta) = V(s;\theta, \beta) + \Bigr( A(s,a;\theta, \alpha) - \frac{1}{|\mathcal{A}|}\sum_{a'}A(s,a';\theta,\alpha) \Bigl),
% \end{equation}
% where $\alpha$ and $\beta$ are the parameters of the two fully connected layers separated for $V$ and $A$.

\subsection{Echo state network}
In this study, a special RNN called echo state network (ESN) is used. ESN has a randomly connected recurrent layer called reservoir and modifies only the weight values for the output layer called readout units, while the other weight values are fixed. The reservoir transforms the given time-series input into a spatio-temporal pattern and generates a state that reflects the past to present memory. It is shown that ESN can learn time-series processing simply by modifying readout units that process the output of the reservoir obtained by nonlinear transformation of reservoir state.

\begin{figure}
\centering
\includegraphics[width=10.0cm]{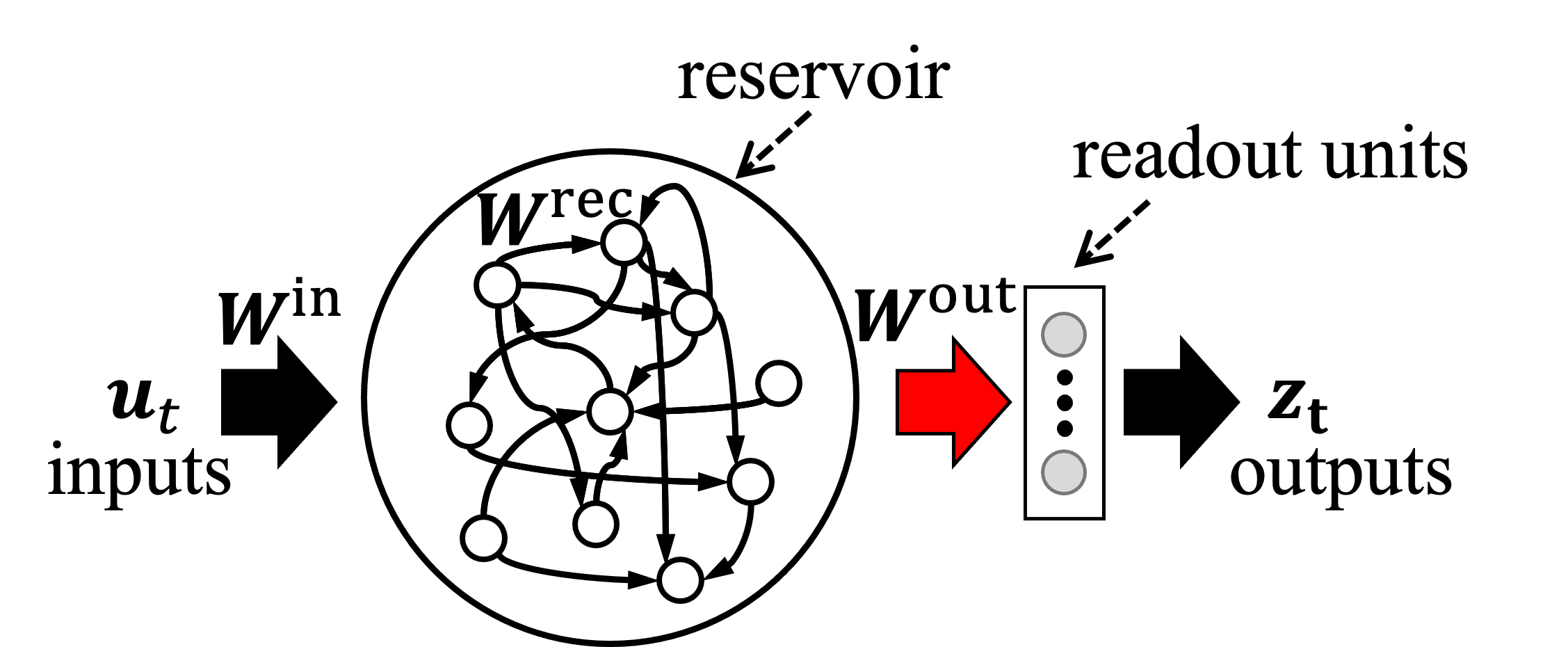}
\caption{Echo state network (ESN).
An ESN is a special recurrent neural network whose recurrent layer is called reservoir. The output layer (readout units) is only trained, while the other weights are fixed.
}
\label{fig:ESN}
\end{figure}

The ESN is shown in Fig. \ref{fig:ESN}. The reservoir consists of $N_{\rm{x}}$ neurons, which are recurrently connected with weight matrix $\bm{W}^{\rm{rec}}$ with connection
probability $p$. The external input is given as a vector
$\bm{u}_t \in \mathbb{R}^{N_i}$, 
and the input is received through weight matrix $\bm{W}^{\rm{in}} \in \mathbb{R}^{N_x \times N_i}$. 
The output of the reservoir neurons is represented by the following equation
\begin{equation}
\label{eq:reservoir_output}
    \bm{x}_t = f(g \bm{W}^{\rm{rec}} \bm{x}_{t-1} + \bm{W}^{\rm{in}}\bm{u}_t + \bm{b}),
\end{equation}
where $g$ is a scaling parameter for the recurrent connection weight matrix and $\bm{b} \in \mathbb{R}^{N_x}$ is bias vector. The activation function of the reservoir neurons is $f(\cdot)=\tanh(\cdot)$.
In ESN, $\bm{W}^{\rm{rec}}$ is one of the most important factors. $\bm{W}^{\rm{rec}}$ is initialized to have a spectral radius of 1 by following procedure. An $N_{\rm{x}} \times N_{\rm{x}}$ matrix $\bm{W}$ is generated randomly (by uniform distribution from $-1$ to $1$ in this study), and then some elements of the matrix are set to $0$ with probability $(1-p)$ to make the connection sparse. The spectral radius of the matrix $\rho (\bm{W})$ is calculated, and the $\bm{W}^{\rm{rec}}$ is determined as follows
\begin{equation}
\label{eq:set_spectrum_radius_1}
    \bm{W}^{\rm{rec}} = \frac{1}{\rho (\bm{W})}\bm{W}.
\end{equation}
With this procedure, $\bm{W}^{\rm{rec}}$ becomes a random sparse matrix with a spectral radius of $1$. As a result, arbitrary constant $g$ can make the spectral radius of $\bm{W}^{\rm{rec}}$ into $g$. In general, $g$ must be set to $g<1$ to store memory of input sequence. A small $g$ tends to cause a rapid loss of memory information, while a large $g$ tends to cause a slowly loss of memory. However, there are many cases where $g>1$ is used, and it is argued that the best value for learning is when the dynamics of the reservoir is around the edge of chaos \cite{bertschinger2004real, Asada}.

\subsection{Deep echo state Q-network}
\begin{figure}
\centering
\includegraphics[width=12.0cm]{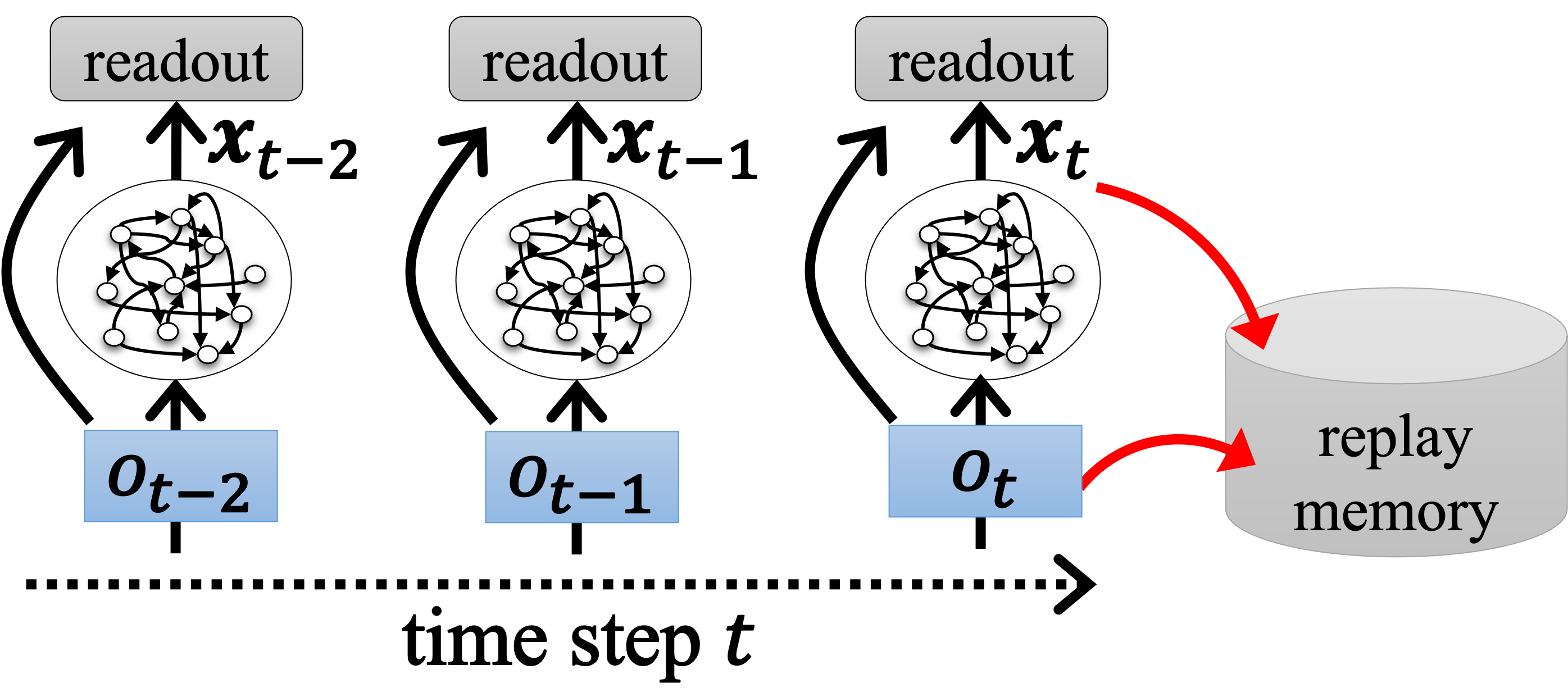}
\caption{Replay memory with reservoir network. In this technique, the observation $\bm{o}_t$ received from the environment is input to the reservoir and the reservoir output $\bm{x}_t$ and $\bm{o}_t$ are stored. The NN learns using the output of the reservoir as input.}
\label{fig:ESRM}
\end{figure}

This study uses deep echo state Q-network \cite{chang2020deep} in which the agent stores the observation $o$ and the reservoir output $\bm{x}$ including the memory of the context of the observation. Fig. \ref{fig:ESRM} shows the conceptual diagram of this method. The state $s_t$ or observation $o_t$ is stored in regular experience replay. Meanwhile, in this method, the $o_t$ is input to the reservoir, and the resulting spatio-temporal pattern output $\bm{x}_t$ and $o_t$ is stored, therefore the experience is stored as $e_k=(o_t, \bm{x}_t, a_t, r_t, o_{t+1}, \bm{x}_{t+1})$ into $\mathcal{D}$. In this way, the experience in the replay memory does not contain only instantaneous information at each step but includes the context of the observation in the reservoir output, then learning can be performed without ignoring the context by simple random sampling. Therefore, RNN can be trained with a simple sampling method without using a complicated procedure such as producing a start state of the RNN by unrolling the network with replay sequence in the past several steps.

\subsection{Multi-layered readout}
The basic ESN model has a single-layered units that performs a linear transformation as an output layer called readout. However, this study uses a multi-layered readout that consists of multi-layered NN. Using multi-layered readout is not common because it often gives inferior results, although a deeply layered NN is more powerful and expressive to map from input to output than a linear layer \cite{lukovsevivcius2009reservoir}. On the other hand, several studies have shown that multi-layered readout is effective when ESN is used for RL \cite{bush2008modeling, matsuki2017reinforcement}. Therefore, this study investigates how the use of multi-layered readouts affects the performance of deep echo state Q-network. The observation $o_t$ is expanded to reservoir output $\bm{x}_t$, therefore the value function network, which is the multi-layered readout, is described as $Q(\bm{o}_t, \bm{x}_t, a_t)$.

\section{Tasks}
\label{sec:task}
The effectiveness of the method is tested using the classic control tasks CartPole, MountainCar, Acrobot, and Pendulum included in the OpenAI Gym. The learning agent observes the states such as position, angle, velocity, and angular velocity from the environment in these tasks. To make these tasks require time-series processing, the tasks are modified so that velocity and angular velocity are unavailable in the experiment. The observed inputs from the environment were normalized so that the range was $-1$ to $1$. Reward values were limited to $-1$, $0$, or $1$ to implement Reward Clipping. The success condition for learning each task is to complete the task in $10$ consecutive trials.

\subsection{CarPole task}
The CartPole is a classical inverted pendulum task that requires the agent to keep a mounted pole within a certain angular range for 200 steps by moving the cart from side to side. This environment corresponds to the task used in the work of Barto et al. \cite{barto1983neuronlike}. According to the source code, the cart position $s^1_t$ is defined in the range of $-4.8$ to $4.8$, and the  pole angle $s^2_t$ is defined in the range of $-0.418$ to $0.418$, therefore the observation at time $t$ is set to $o_t = \{s^1_t / 4.8, s^2_t / 0.418 \}$. An episode ends when the cart position or the pole angle exceeds the certain range, or when 200 steps have elapsed. In this experiment, the agent receives a reward $r_t = 1$ at the end of the episode if the step for which the pole has stood on exceeds $195$, and a reward $r_t=-1$ otherwise. For all other steps, $r_t = 0$.

\subsection{MountainCar task}
The MountainCar is a task in which an agent moves a car from side to side on a one-dimensional course between two mountains to reach the goal set on top of the right mountain. The agent have to gain momentum by driving the car back and forth between the mountains to complete the task. This task was first described in Moore's research \cite{moore1990efficient}. According to the source code, the car's position $s^1_t$ is defined in the range of $-1.2$ to $0.6$, therefore the observation is set to $o_t=\big\{(s^1_t+0.3)/0.9\big\}$. An episode ends when the car reaches the goal or $200$ steps have elapsed. In this experiment, the agent receives a reward $r_t=1$ when it reaches the goal and $r_t=-1$ for other steps.

\subsection{Acrobot task}
The Acrobot is a double pendulum swing-up task, where the goal is to swing the tip above a specified height by selecting the direction of a constant torque applied to the central link. To complete the task, the agent must manipulate the link well to gain momentum. This task was first described by Sutton \cite{sutton1996generalization}. According to the source code, the state is defined as $s^1_t=\cos{\theta^1_t}, s^2_t=\sin{\theta^1_t}, s^3_t=\cos{\theta^2_t}, s^4_t=\sin{\theta^2_t}$, where $\theta^1_t$ is the angle of the first link from the axis in the vertical downward direction and $\theta^2_t$ is the relative angle of the two links at time $t$. The observation is defined as $o_t=\{s^1_t, s^2_t, s^3_t, s^4_t\}$. An episode ends when the tip exceeds the specified height, or $200$ steps have elapsed. In this experiment, the agent receives a reward $r_t=1$ when the tip exceeds the specified height and receives a reward $r_t=-1$ for all other steps.

\subsection{Pendulum task}
The Pendulum is a task in which the agent swing up a pendulum by determining the torque $\tau$ applied to a single-link pendulum. The torque that the agent can apply is not enough to swing the pendulum up; therefore, the agent must swing it from side to side to give it momentum. According to the source code, the state is defined as
$s^1_t=\cos{\theta_t},s^2_t=\sin{\theta_t}$ based on the angle $\theta_t$ from the upward direction at time $t$. Here, $\theta$ is normalized in the range of $-\pi$ to $\pi$. The observation is defined as $o_t=\{s^1_t,s^2_t\}$. An episode ends when $200$ steps have passed since the start. 
This experiment focuses on learning discrete action selection; therefore, the agent chooses an action to set the torque to either $-1$ or $1$.
A reward implemented in the library is $R_t=-\theta^2-0.1 \Dot{\theta}-0.0012 \tau^2$.
Then the reward is set to $r_t=-1$, when $R_t \leq -1$ and $r_t=1$ when $R_t > -1$.

\section{Experimental results}
\label{sec:result}
\subsection{Common hyper parameters}
\label{subsec:common_hyper_parameters}
The common parameters in the experiments are set as follows. The neurons in the reservoir connect recurrently with a connection probability $p=0.1$. $g$ is set to $0.9$ and the input weight matrix $\bm{W}^{\rm{in}}$ is generated from a uniform distribution between $-1$ and $1$. The bias vector $\bm{b}$ is generated from a uniform distribution between -0.2 and 0.2. A multi-layered readout is a neural network having a hidden layer that consists of $250$ neurons whose activation function is ReLu. Training is performed using AMSGrad with a momentum of $0.9$. The learning rate is $0.001$ for CartPole, Acrobot and Pendulum, and $0.005$ for MountainCar. $N_D = 256$ experiences are randomly sampled from $\mathcal{D}$ of maximum size $N=10000$. 
See Appendix for an investigation to determine the value of the learning rates. 
Once every two episodes, the parameters $\theta_i$ of the main network is copied to the parameters $\theta^-$ of the target network. The discount rate $\gamma$ is $0.99$. $\epsilon$ is set to 0.5 at the beginning of training and decays every trial by multiplying $\sqrt[400]{0.02}$ to become $0.01$ until the $401$th episode. When the agent complete the task 10 times in a row or the number of trials reaches $500$, learning is terminated. The number of the reservoir neurons is $N_x=50$ for each task.

\subsection{Training results}
\label{subsec:training_result}
\begin{figure}
  \begin{minipage}[b]{0.45\linewidth}
    \centering
    \includegraphics[keepaspectratio, scale=0.5]{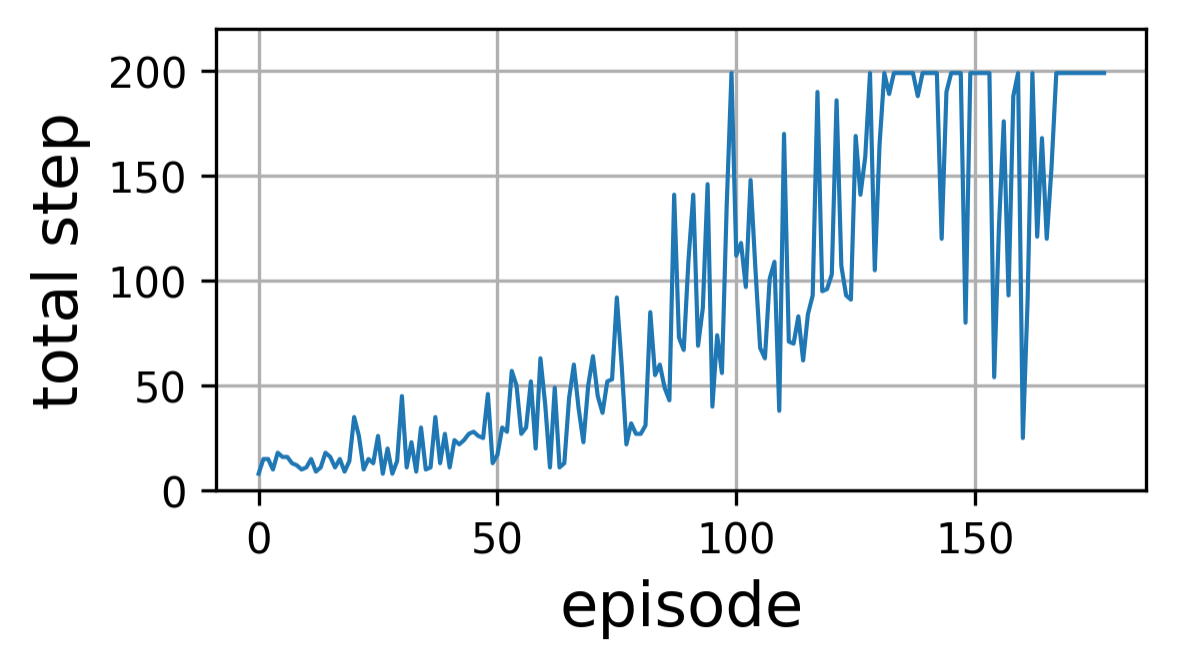}
    \subcaption{CartPole}
  \end{minipage}
  \begin{minipage}[b]{0.45\linewidth}
    \centering
    \includegraphics[keepaspectratio, scale=0.5]{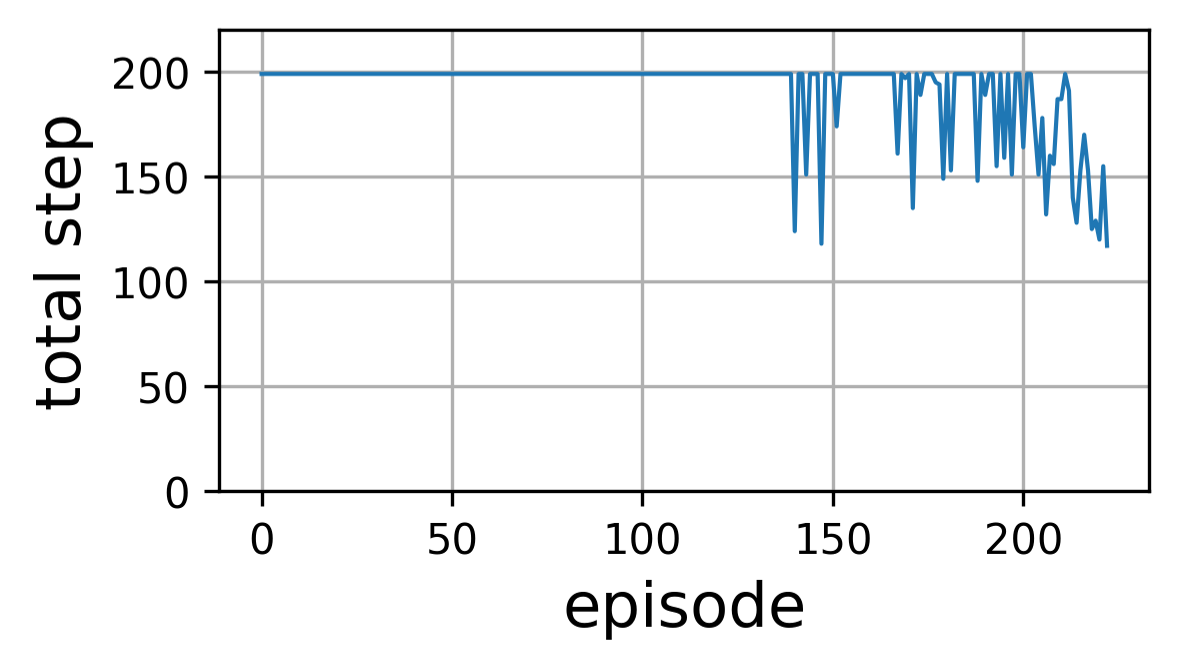}
    \subcaption{MountainCar}
  \end{minipage}
\\
  \begin{minipage}[b]{0.45\linewidth}
    \centering
    \includegraphics[keepaspectratio, scale=0.5]{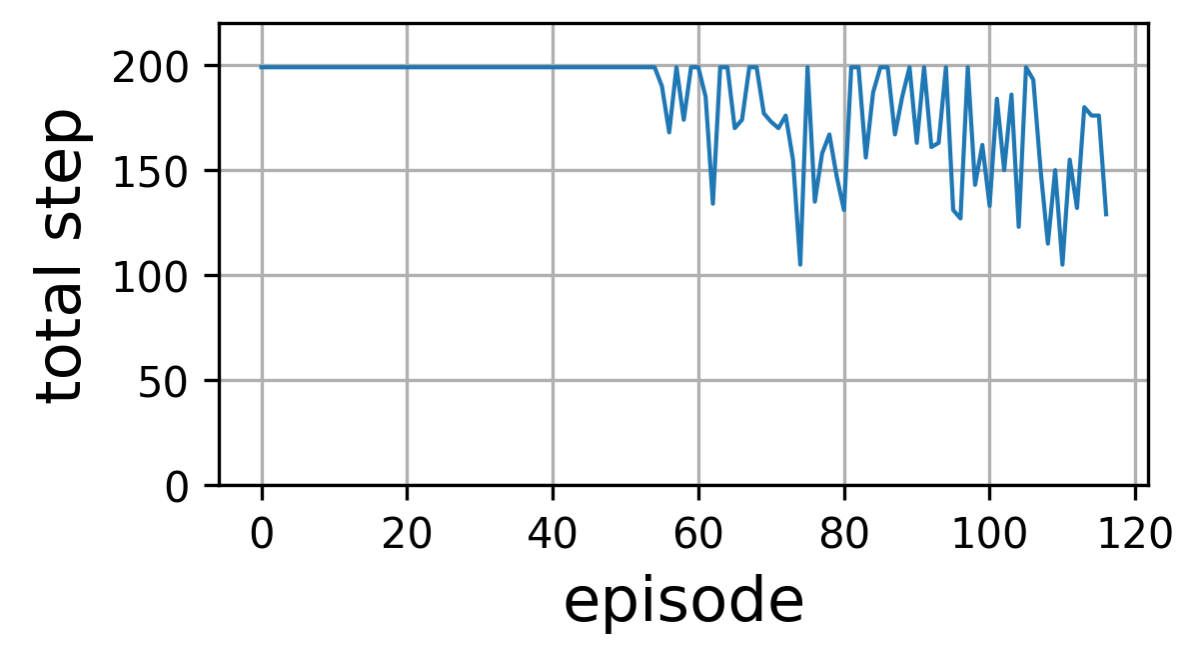}
    \subcaption{Acrobot}
  \end{minipage}
  \begin{minipage}[b]{0.45\linewidth}
    \centering
    \includegraphics[keepaspectratio, scale=0.5]{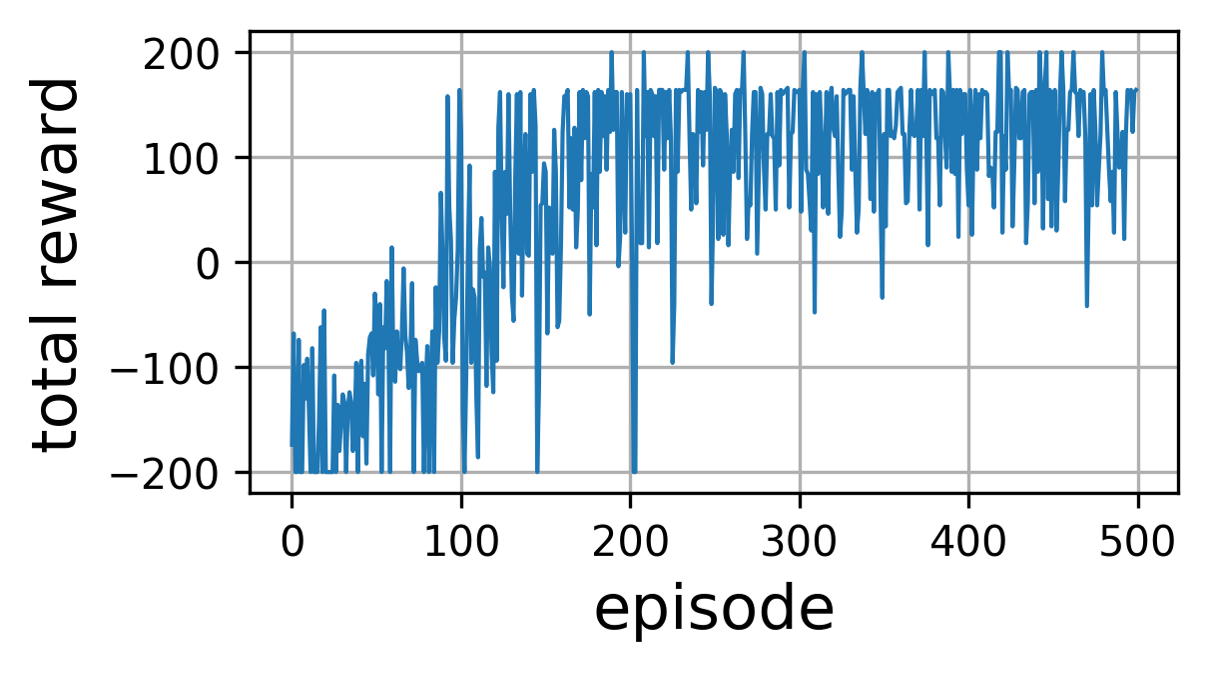}
    \subcaption{Pendulum}
  \end{minipage}
  \caption{Learning curves of each task CartPole (a), MountainCar (b), Actobot (c), and Pendulum (d). The horizontal axis indicates the number of episodes. The vertical axis shows the number of steps in each episode in (a-c) and the total reward in (d).
}
\label{fig:LC}
\end{figure}

\begin{figure}[htbp]
  \begin{minipage}[b]{0.45\linewidth}
    \centering
    \includegraphics[keepaspectratio, scale=0.5]{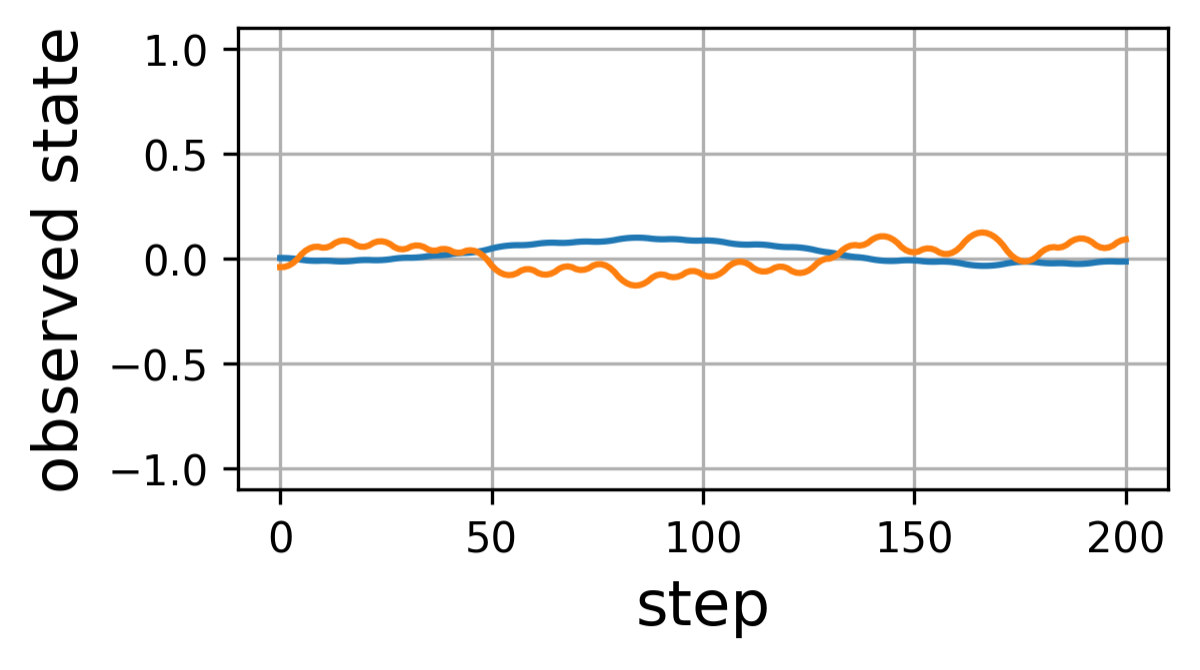}
    \subcaption{CartPole}
  \end{minipage}
  \begin{minipage}[b]{0.45\linewidth}
    \centering
    \includegraphics[keepaspectratio, scale=0.5]{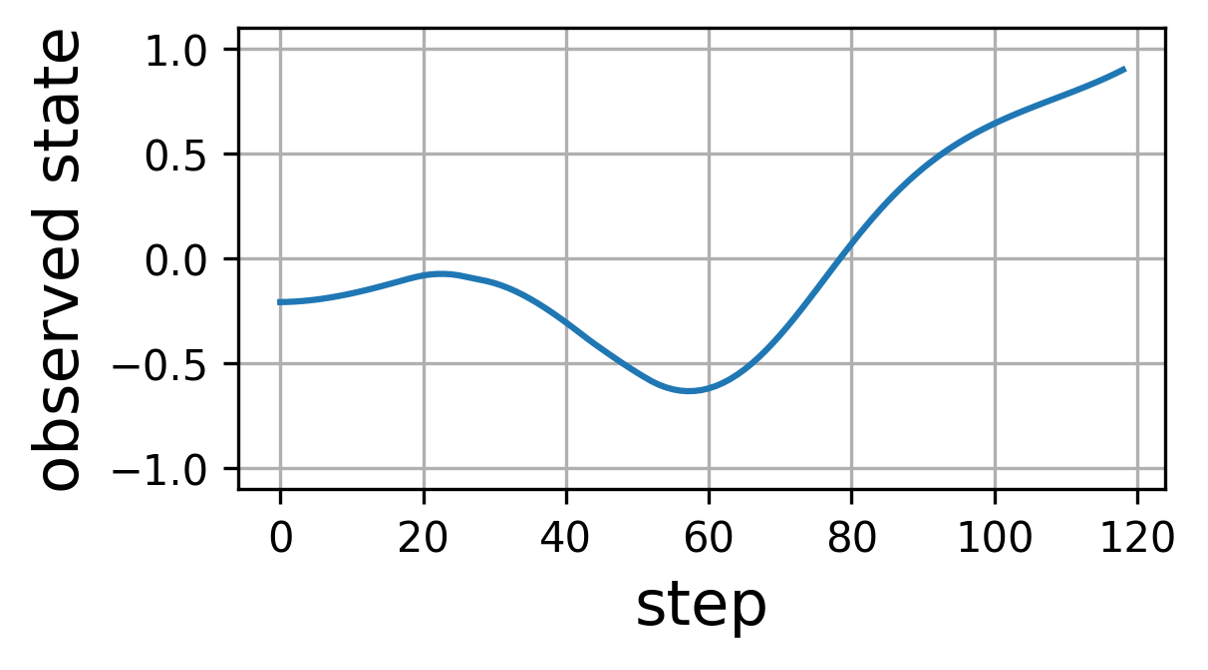}
    \subcaption{MountainCar}
  \end{minipage}
\\
  \begin{minipage}[b]{0.45\linewidth}
    \centering
    \includegraphics[keepaspectratio, scale=0.5]{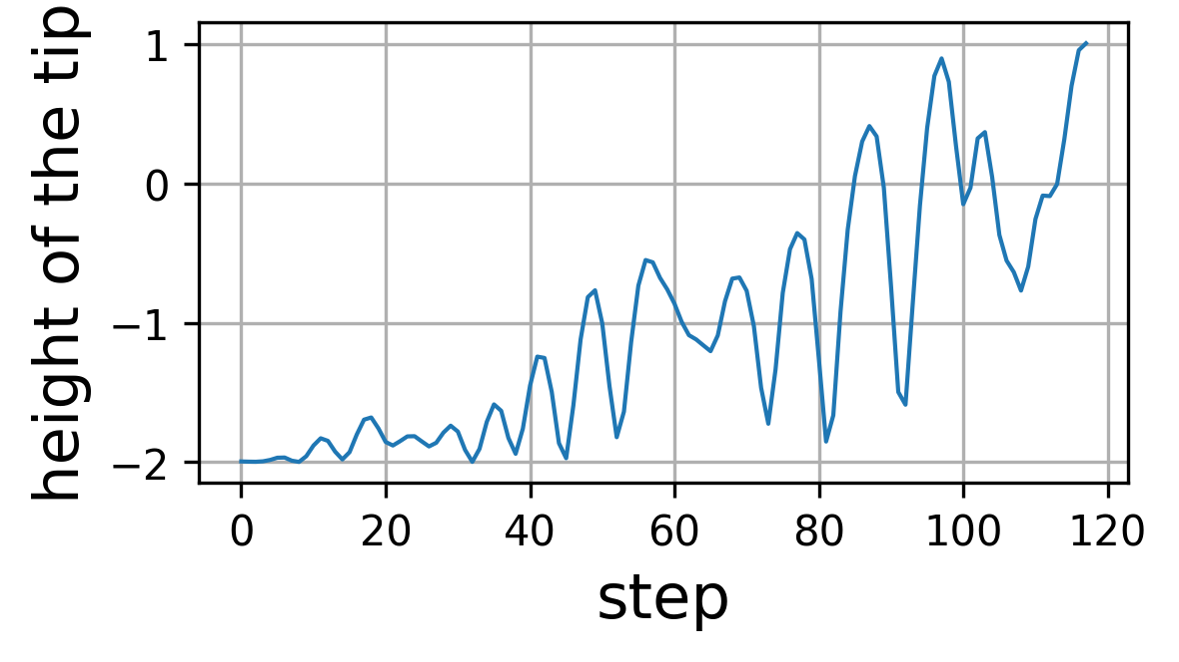}
    \subcaption{Acrobot}
  \end{minipage}
  \begin{minipage}[b]{0.45\linewidth}
    \centering
    \includegraphics[keepaspectratio, scale=0.5]{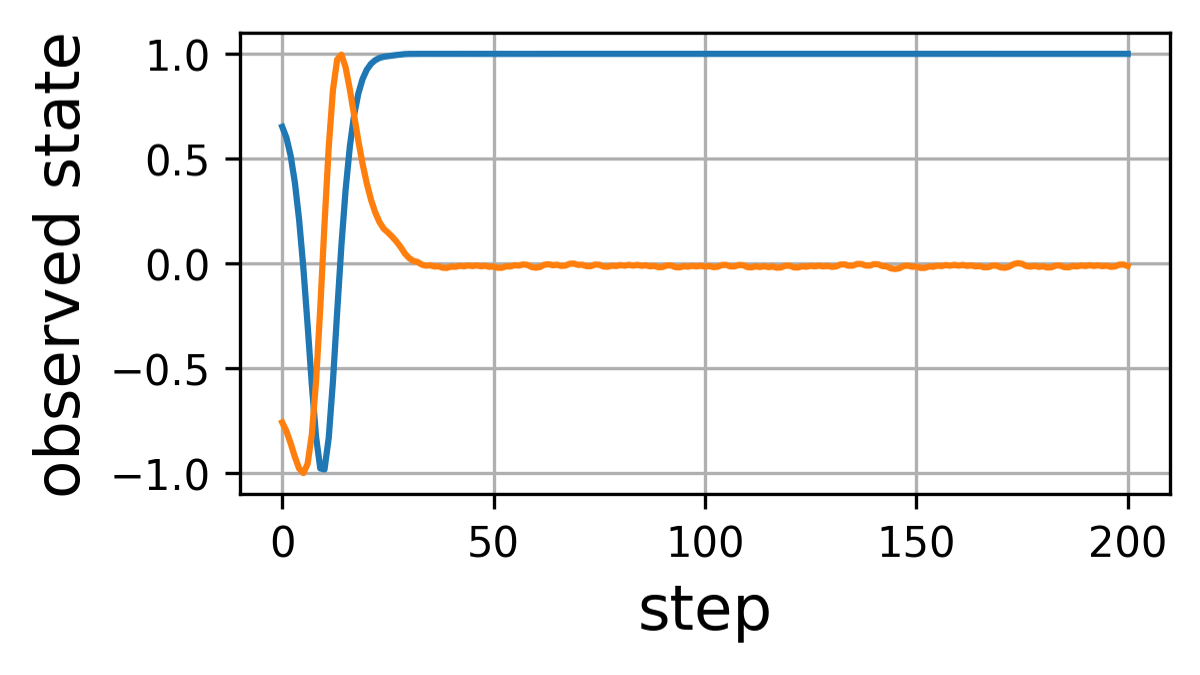}
    \subcaption{Pendulum}
  \end{minipage}
  \caption{State of the agent in each task. (a) blue line shows the position of the cart, orange line shows the angle of the pole. (b) blue line indicates the position of the car. (c) blue line shows the height of the tip of the double pendulum. (d) blue line shows $\cos{\theta}$, orange line shows $\sin{\theta}$, where $\theta$ is the angle of the pendulum.
}
\label{fig:OBS}
\end{figure}

Learning are conducted with four different tasks to confirm whether the agent can learn tasks that require time-series processing. 
Fig. \ref{fig:LC} shows the learning curve in each task. Fig. \ref{fig:OBS} shows the state of the agent in a test trial without learning and exploration after the training. Note that, in Fig. \ref{fig:OBS} (a), (b), and (d) show the observation of the agent, while only (c) shows the height of the tip of the double pendulum for visibility of the results.

The result in the CartPole task is shown in Fig. \ref{fig:LC} (a), where the agent completed the task when the number of steps in an episode reaches $200$. The figure shows that the agent learned as the episode progresses, and the number of steps the pole stands increases. In the $178$th episode, the agent completed the task $10$ times in a row.
Fig. \ref{fig:OBS} (a) shows that the cart is stable near the center, and the pole angle is kept near $0$ while swinging slightly. This result indicates that the agent successfully made the pole stand. The result indicates that the context of the observation is retained as short-term memory in the reservoir dynamics and the agent successfully learned by observing only the current cart position and pole angle without velocity and angular velocity.

The result in the  MountainCar task is shown in Fig. \ref{fig:LC} (b), where the agent reaches the goal when the number of steps in an episode is less than $200$. From this figure, we can see that the agent learned and became able to reach the goal on the mountain. In the $223$th episode, the agent completed the task $10$ times in a row.
Fig. \ref{fig:OBS} (b) shows that the car's position dropped once to around -0.6 and then rose, and the agent completed the task in $119$ steps. This result indicates that the agent successfully learned to back the car once, climb up a small mountain behind it, and then run-up to the goal using the momentum of the descent. To learn this task statically, the agent needs to observe the velocity of the car to distinguish if the car is climbing or descending the mountain. The result indicates that it has successfully learned to process the time-series of position to discriminate between climbing and descending.

Fig. \ref{fig:LC} (c) shows the learning curve of the Acrobot task, where the agent completes the task when the number of steps in an episode is less than $200$. This figure shows that the agent learned as the episode progresses and swung the tip higher than the specified height. In the $117$th episode, the agent succeeded in reaching the goal $10$ times in a row. 
Fig. \ref{fig:OBS} (c) shows that the tip of the double pendulum gradually increased in height while oscillating and successfully swung up after $118$ steps. This result indicates that the agent successfully learned to swing the pendulum by applying appropriate torque and swinging up the tip using inertia. In this task, it is essential to know which direction of inertia the arm has at each time to control it properly. The agent successfully learned to control the pendulum without observing the angular velocity, then this result indicates that the agent can determine the direction of inertia from the series of angle observation at each time.

The result in the Pendulum task is shown in Fig. \ref{fig:LC} (d).
This figure shows that the total reward increases as the episode progresses, indicating that the agent learned to swing up the pendulum more quickly. This task has no explicit goal; therefore, the learning continues up to the upper limit of $500$ episodes.
Fig. \ref{fig:OBS} (d) shows that the agent successfully swung the pendulum up in about $30$ steps, gaining momentum while shaking the pendulum, and then successfully maintained the state. This result indicates that the agent successfully learned to swing the pendulum up by applying torque in the appropriate direction and using inertia without observing angular velocity at each time.

These results indicate that deep echo state Q-network successfully works to learn control tasks that require time-series processing.

\subsection{Relationship between spectral radius and learning performance}
\label{subsec:g_and_performance}
\begin{figure}[t]
  \begin{minipage}[b]{0.45\linewidth}
    \centering
    \includegraphics[keepaspectratio, scale=0.3]{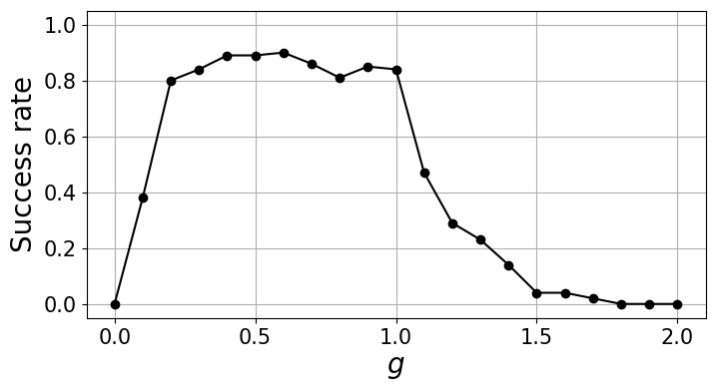}
    \subcaption{CartPole}
  \end{minipage}
  \begin{minipage}[b]{0.45\linewidth}
    \centering
    \includegraphics[keepaspectratio, scale=0.3]{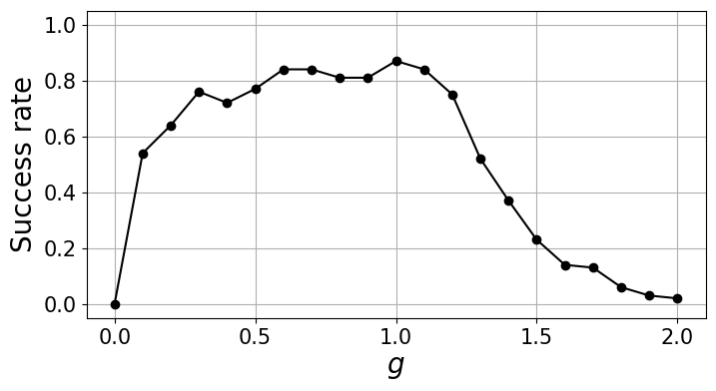}
    \subcaption{MountainCar}
  \end{minipage}
\\
  \begin{minipage}[b]{0.45\linewidth}
    \centering
    \includegraphics[keepaspectratio, scale=0.3]{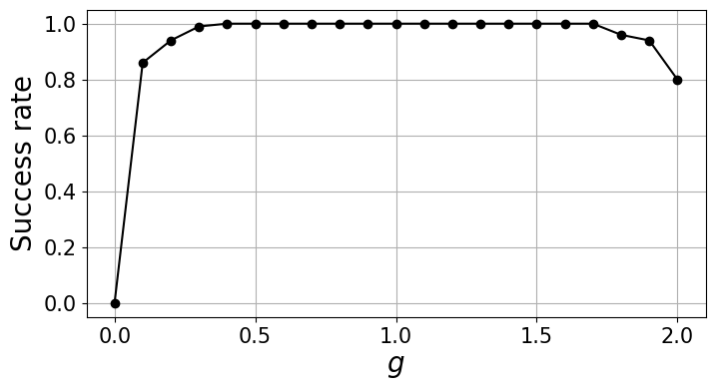}
    \subcaption{Acrobot}
  \end{minipage}
  \begin{minipage}[b]{0.45\linewidth}
    \centering
    \includegraphics[keepaspectratio, scale=0.3]{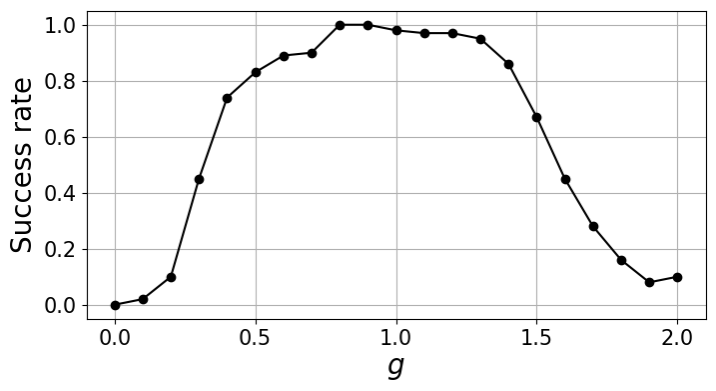}
    \subcaption{Pendulum}
  \end{minipage}
  \caption{The success rate for each $g$. The values are calculated with $100$ trials with a different random number sequence. The vertical axis indicates the success rate, and the horizontal axis indicates $g$, which determines the spectral radius of $\bm{W}^{\rm{rec}}$.
}
\label{fig:SR}
\end{figure}

The parameter $g$ that adjusts the memory length of the reservoir and changes its dynamics is a very important parameter because the adjustment of this value strongly affects the learning performance of the ESN.
To examine the effect of the reservoir dynamics on the success or failure of learning, $g$ was varied from 0 to 2 in increments of 0.1 and examined the rate of successful learning cases among $100$ random sequences. Here, successful learning is defined as achieving the goal required by the task for $10$ consecutive episodes. For the Pendulum task, success condition is defined as the agent kept swinging for the last $50$ steps in one episode. Fig. \ref{fig:SR} shows the results. The common trend for all tasks is that the agent failed when the spectral radius was too small or large and always showed high performance when the spectral radius is slightly less than $1$. This result is consistent with the general knowledge in the study of reservoir computing. The range of $g$ showing the high performance varies depending on the task. It can be considered that the difference in input dimension, dynamics, or the required memory length of each task affects the appropriate value of $g$. When the spectral radius is zero, the success rate is zero in all tasks. This indicates that the memory of the observation held in the reservoir plays an essential role in learning the tasks that require time-series processing.

\subsection{Learning performance with a single layer readout}
\begin{figure}[htbp]
  \begin{minipage}[b]{0.45\linewidth}
    \centering
    \includegraphics[keepaspectratio, scale=0.3]{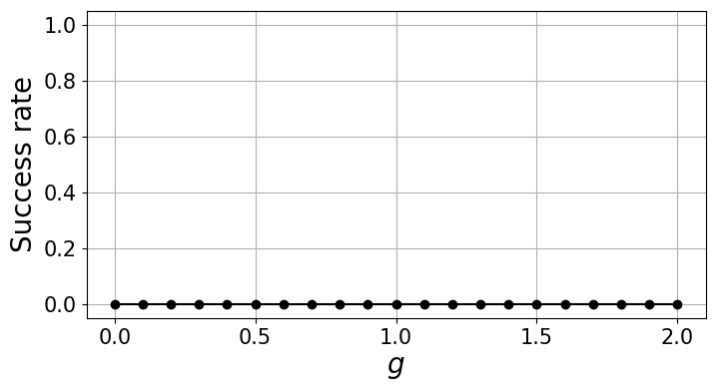}
    \subcaption{CartPole}
  \end{minipage}
  \begin{minipage}[b]{0.45\linewidth}
    \centering
    \includegraphics[keepaspectratio, scale=0.3]{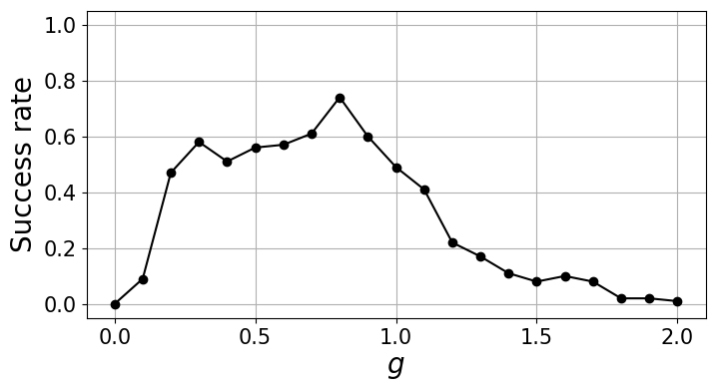}
    \subcaption{MountainCar}
  \end{minipage}
\\
  \begin{minipage}[b]{0.45\linewidth}
    \centering
    \includegraphics[keepaspectratio, scale=0.3]{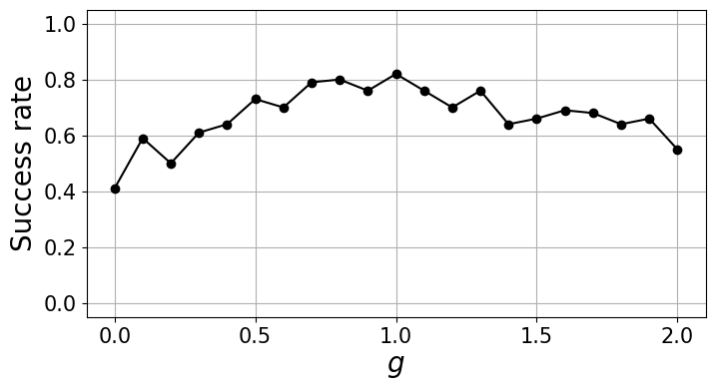}
    \subcaption{Acrobot}
  \end{minipage}
  \begin{minipage}[b]{0.45\linewidth}
    \centering
    \includegraphics[keepaspectratio, scale=0.3]{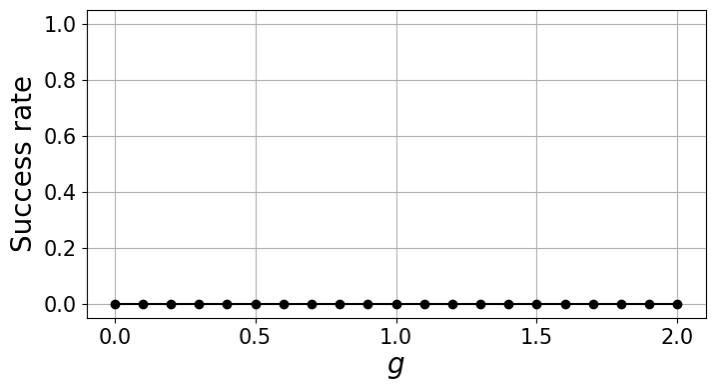}
    \subcaption{Pendulum}
  \end{minipage}
  \caption{The success rate with single layer readout for each $g$. The values are calculated with $100$ trials with a different random number sequence. The vertical axis indicates the success rate, and the horizontal axis indicates $g$, which determines the spectral radius of $\bm{W}^{\rm{rec}}$.
}
\label{fig:SR_SINGLE}
\end{figure}

The learning performance was compared with an agent having a single-layered readout trained with the same tasks to confirm the effectiveness of using a multi-layered readout. 
In this experiment, the readout consisted of a linear layer that is commonly used in the reservoir computing field, and the parameters other than the learning rate for MountainCar were the same as described in Section \ref{subsec:common_hyper_parameters}.
The learning rate for MountainCar was set to 0.01.
See Appendix for an investigation to determine the value of the learning rates. 
In the same way as in Section \ref{subsec:g_and_performance}, the spectral radius was varied and examined the success rate with $100$ random sequences.
The results are shown in Fig. \ref{fig:SR_SINGLE}. This figure shows that the learning performance of MountainCar and Acrobot deteriorated, and the agent failed to learn CartPole and Pendulum tasks when using a single layer readout.
These results indicate that using a multi-layered readout improves the performance of deep echo state Q-network in the whole range of the parameter $g$.

\section{Conclusion}
\label{sec:conclusion}
 Various methods have been developed to train RNN with RL for time-dependent tasks. Although they have shown powerful performance, the procedures for experience replay tend to be more complicated and computationally expensive and the vanishing/exploding gradients problem of the RNN still remains. On the other hand, deep echo state Q-network in which the system stores the reservoir output containing memory information into a replay memory, eliminates such issues and realizes learning of tasks that require time-series processing using only a simple sampling technique.
 
 This study shows that introducing multi-layered readout into the deep echo state Q-network approach improves the learning performance of control tasks that require time-series processing. This result is one example that demonstrates multi-layered readout is effective in training ESN by RL.

DRL techniques have evolved by influencing each other and merging, and now there are a huge variety of implementations. Introducing reservoir to replay memory is a simple technique in which the observation is filtered once by the reservoir network. Therefore, any DRL methods can combine this approach as long as it uses experience replay. Just as using multi-layered neural networks improved the learning performance in this study, it can be expected that this method will handle more difficult time-series tasks when combined with various deep learning and DRL methods.

However, there is a clear and hard challenge to be overcome to use a reservoir to learn more complex tasks with RL. The challenge is that reservoir networks have difficulty dealing with high-dimensional inputs such as images. Although the number of neurons in the reservoir must be sufficiently larger than the input dimension, increasing the number of reservoir neurons causes an increase in computational cost or overfitting the model. Tong and Tanaka showed that reservoirs can process time-series of image input using untrained CNN \cite{tong2018reservoir}, and Chang and Futagami combined this approach with evolutionary computation techniques to successfully learn a car racing game with raw pixel inputs \cite{chang2020reinforcement}. It may be possible to overcome the challenges of dealing with high-dimensional inputs by using other techniques to extract features and then using the reservoir network specifically for learning policy.

In addition, various techniques in reservoir computing may be introduced to enhance the capability of the proposed method. For example, Deep ESN and parallel ESN have been proposed in which multiple reservoirs are configured in multi layer or parallel \cite{gallicchio2017deep, gallicchio2017echo, ma2017deep}. Sakemi et al. succeeded in maintaining performance with fewer reservoir neurons by using a structure that sends the reservoir output of several steps to the readout \cite{sakemi2020model}. The future work is to improve the performance of RL using reservoir by introducing the above techniques and many other reservoir computing knowledge.

%エッジ学習のことも触れるか

\section*{Acknowledgments}
The author would like to thank Prof. Katsunari Shibata for useful discussions about this research.
This work was supported by JSPS KAKENHI (Grant-in-Aid for Encouragement of Scientists) Number 21H04323. 

\appendix
\section{varying the learning rate and selecting optimizer}
\begin{figure}[htbp]
  \begin{minipage}[b]{0.45\linewidth}
    \centering
    \includegraphics[keepaspectratio, scale=0.3]{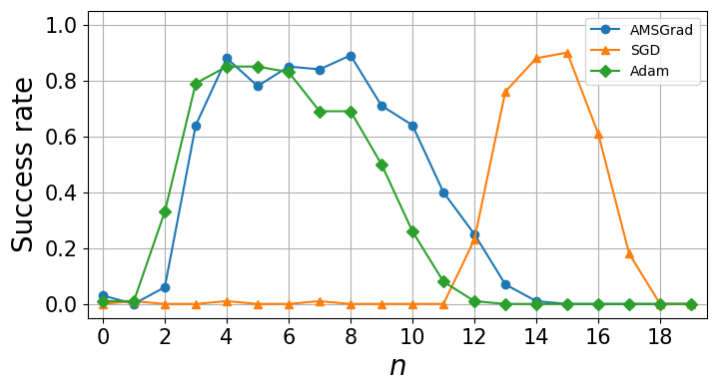}
    \subcaption{CartPole}
  \end{minipage}
  \begin{minipage}[b]{0.45\linewidth}
    \centering
    \includegraphics[keepaspectratio, scale=0.3]{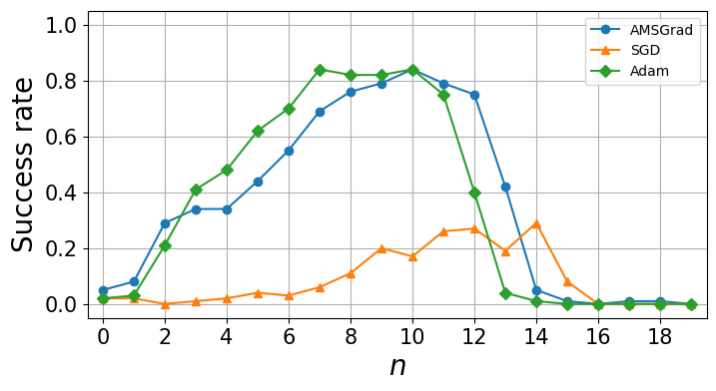}
    \subcaption{MountainCar}
  \end{minipage}
\\
  \begin{minipage}[b]{0.45\linewidth}
    \centering
    \includegraphics[keepaspectratio, scale=0.3]{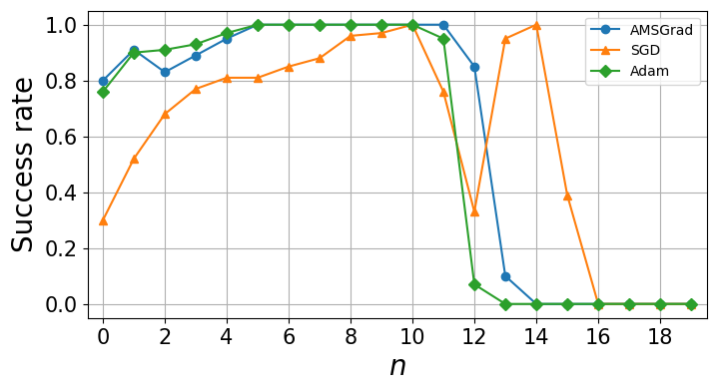}
    \subcaption{Acrobot}
  \end{minipage}
  \begin{minipage}[b]{0.45\linewidth}
    \centering
    \includegraphics[keepaspectratio, scale=0.3]{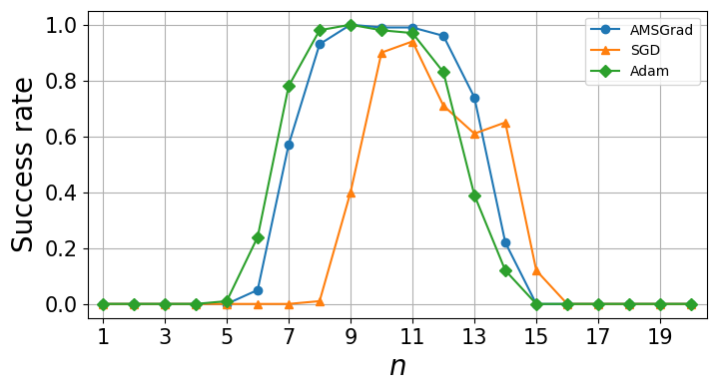}
    \subcaption{Pendulum}
  \end{minipage}
  \caption{The learning rate for each optimizer and varied learning rate of the agent with the multi-layered readout. The values are calculated with $100$ trials with a different random number sequence. The vertical axis indicates the success rate, and the horizontal axis indicates $n$, which determines the learning rate.
}
\label{fig:LR}
\end{figure}

The choice of optimizer and the setting of learning rate strongly influence the learning of a neural network.
To determine the parameters to be used in the experiment, we investigated the change in the learning success rate while varying learning rate for each AMSGrad, SGD, and Adam, here the learning rate was set to $0.000005 \times 2^n$ and $n$ was varied from 0 to 19.
Fig. \ref{fig:LR} shows the results.
From this Figure, it can be seen that there is no significant difference between AMSGrad and Adam, but overall the performance of the other optimizers is higher than that of SGD. In this study, AMSGrad was used and learning rate was determined based on the result in this verification.

\begin{figure}[htbp]
  \begin{minipage}[b]{0.45\linewidth}
    \centering
    \includegraphics[keepaspectratio, scale=0.3]{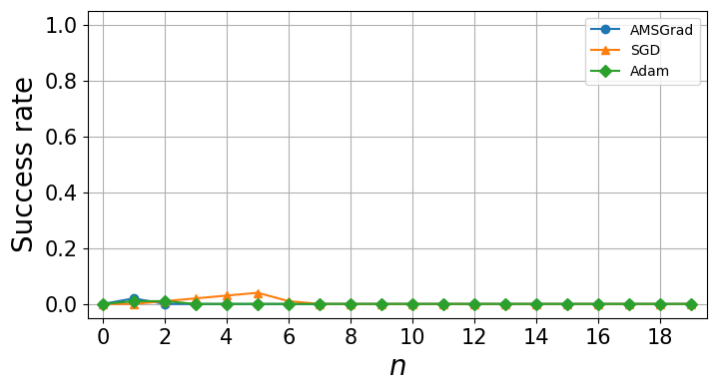}
    \subcaption{CartPole}
  \end{minipage}
  \begin{minipage}[b]{0.45\linewidth}
    \centering
    \includegraphics[keepaspectratio, scale=0.3]{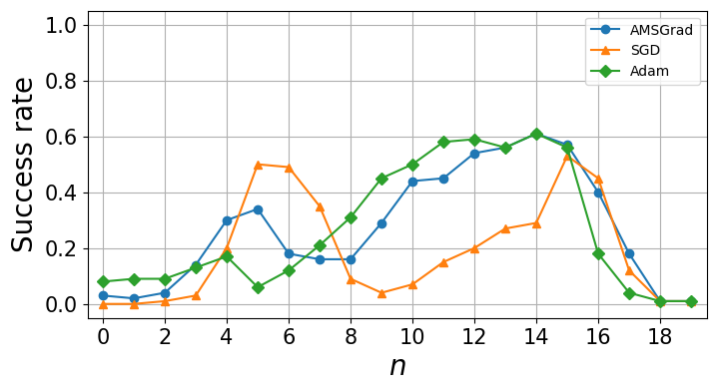}
    \subcaption{MountainCar}
  \end{minipage}
\\
  \begin{minipage}[b]{0.45\linewidth}
    \centering
    \includegraphics[keepaspectratio, scale=0.3]{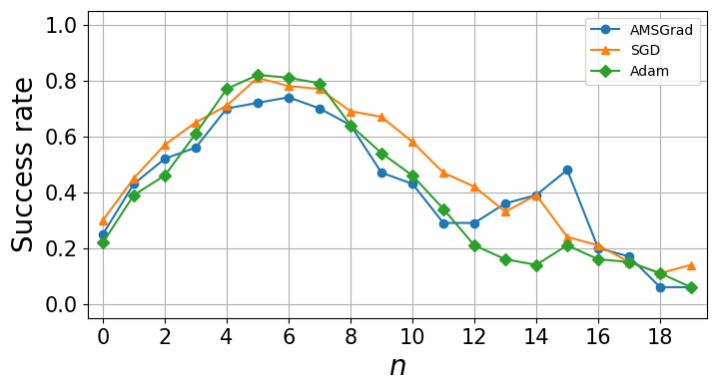}
    \subcaption{Acrobot}
  \end{minipage}
  \begin{minipage}[b]{0.45\linewidth}
    \centering
    \includegraphics[keepaspectratio, scale=0.3]{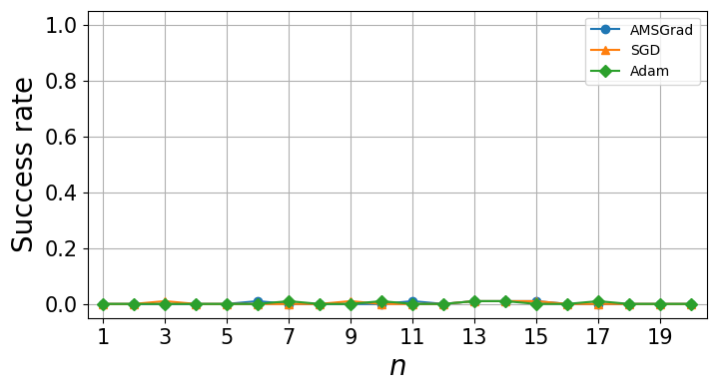}
    \subcaption{Pendulum}
  \end{minipage}
  \caption{The learning rate for each optimizer and varied learning rate of the agent with single-layered readout. The values are calculated with $100$ trials with a different random number sequence. The vertical axis indicates the success rate, and the horizontal axis indicates $n$, which determines the learning rate.
}
\label{fig:LR_SINGLE}
\end{figure}

The same validation was performed for the case of the learning agent with a single-layered readout. Fig. \ref{fig:LR_SINGLE} shows the results that the agent failed to learn CartPole and Pendulum, and demonstrates lower performance in MountainCar and Acrobot than in the case of the multi-layered readout.
These results indicate that the inferior performance of the single-layered readout compared to the multi-layered readout is not due to insufficient optimization of the learning rate and selection of the optimizer. Learning rates in this study for MountainCar and Acrobot were determined based on the result in this verification, and the same value in the case of the multi-layered readout was used for CartPole and Pendulum.

\bibliography{mybibfile}

\end{document}